\documentclass[10pt,conference]{IEEEtran}
\usepackage[tight,footnotesize]{subfigure}
\usepackage{tikz,pgfplots} 
\usepackage{booktabs, multicol, multirow}
\usepackage[acronym,nomain]{glossaries}
\usepackage[inline]{enumitem}
\usepackage{algorithm}
\usepackage{algorithmicx}
\usepackage{algpseudocode}
\usepackage{amsmath,amsfonts,amssymb}
\usepackage[small,bf,hypcap=false]{caption}
\usepackage{subfigure}
\usepackage{grffile}
\usepackage{tabularx}
\usepackage{colortbl}
\usepackage{hhline}
\usepackage{hyperref}
\usepackage[flushleft]{threeparttable}
\usepackage{booktabs,fixltx2e}
\usepackage{balance}
\usepackage{boldline}
\usepackage{rotating}
\usepackage{pdflscape}
\usepackage[many]{tcolorbox}
\usepackage{framed}
\usepackage{cite}
\colorlet{shadecolor}{blue!20}

\renewcommand{\mathbf}[1]{{\boldsymbol #1}}
\hypersetup{
	colorlinks   = true,
	citecolor    = blue,
	linkcolor    = blue,
}
\setlist[itemize]{leftmargin=*}
\setlist[enumerate]{leftmargin=*}

\pgfplotsset{ 
  legend style = {font=\small\sffamily}, 
  label style = {font=\small\sffamily} 
}

\usepackage[colorinlistoftodos,prependcaption,textsize=tiny]{todonotes}
\newcommand{\secref}[1]{Section~\ref{sec:#1}}
\newcommand{\tabref}[1]{Table~\ref{tab:#1}}
\newcommand{\figref}[1]{Figure~\ref{fig:#1}}

\makeglossaries

\begin{document}
%

\title{Transfer Learning for Performance Modeling of Configurable Systems: An Exploratory Analysis}

\author{\IEEEauthorblockN{Pooyan Jamshidi}
\IEEEauthorblockA{
Carnegie Mellon University, USA}
\and
\IEEEauthorblockN{Norbert Siegmund}
\IEEEauthorblockA{
Bauhaus-University Weimar, Germany}
\and
\IEEEauthorblockN{Miguel Velez, Christian K{\"a}stner\\ Akshay Patel, Yuvraj Agarwal}
\IEEEauthorblockA{
Carnegie Mellon University, USA}}

\maketitle
\begin{abstract}

Modern software systems provide many configuration options which significantly influence their non-functional properties. To understand and predict the effect of configuration options, several sampling and learning strategies have been proposed, albeit often with significant cost to cover the highly dimensional configuration space. Recently, transfer learning has been applied to reduce the effort of constructing performance models by transferring knowledge about performance behavior across environments. 
While this line of research is promising to learn more accurate models at a lower cost, it is unclear why and when transfer learning works for performance modeling. To shed light on when it is beneficial to apply transfer learning, we conducted an empirical study on four popular software systems, varying software configurations and environmental conditions, such as hardware, workload, and software versions, to identify the key knowledge pieces that can be exploited for transfer learning. Our results show that in small environmental changes (e.g., homogeneous workload change), by applying a linear transformation to the performance model, we can understand the performance behavior of the target environment, while for severe environmental changes (e.g., drastic workload change) we can transfer only knowledge that makes sampling more efficient, e.g., by reducing the dimensionality of the configuration space.

\end{abstract}


\section{Introduction}
\label{sec:introduction}


Highly configurable software systems, such as mobile apps, compilers, and big data engines, are increasingly exposed to end users and developers on a daily basis for varying use cases. Users are interested not only in the fastest configuration, but also in whether the fastest configuration for their applications also remains the fastest when the environmental situation has been changed. For instance, a mobile developer might be interested to know if the software that she has configured to consume minimal energy on a testing platform will also remain energy efficient on the users' mobile platform; or, in general, whether the configuration will remain optimal when the software is used in a different environment (\emph{e.g.}, with a different \emph{workload}, on different \emph{hardware}). 




\begin{figure}[t]
	\begin{center}
		\includegraphics[width=0.8\columnwidth]{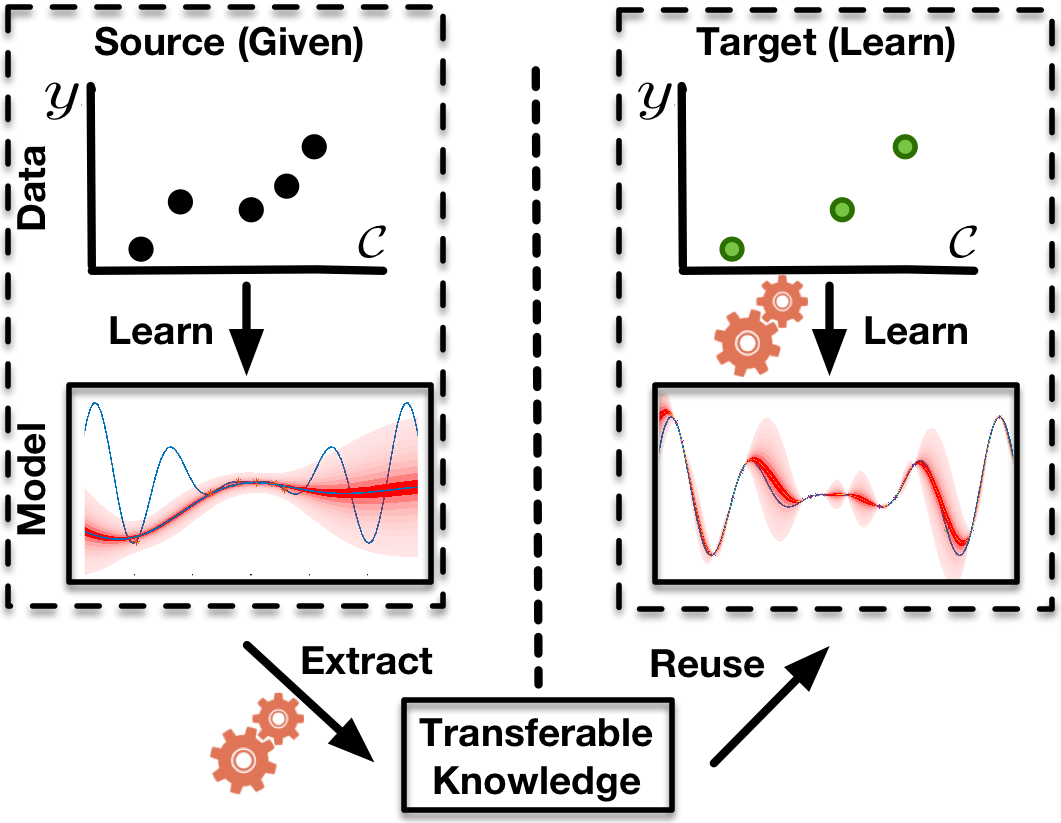}
		\caption{Transfer learning is a form of machine learning that takes advantage of transferable knowledge from source to learn an accurate, reliable, and less costly model for the target environment.}
		\label{fig:tl-general}
	\end{center}
\end{figure}

Performance models have been extensively used to learn and describe the performance behavior of configurable systems~\cite{SGAK:ESECFSE15,SGSAC:ASE15,SKKABRS:ICSE12,JC:MASCOTS16,GCASW:ASE13,YWLE:MASCOTS13,NMSA:Arxive17,ZGBC:ASE15,HSCMAR:SIGPLAN,H:CACM,WWHJK:GECCO15}. However, the exponentially growing configuration space, complex interactions, and unknown constraints among configuration options~\cite{XJFZPT:FSE15} often make it \emph{costly} and difficult to learn an accurate and reliable performance model. Even worse, existing techniques usually consider only a fixed environment (\emph{e.g.}, fixed workload, fixed hardware, fixed versions of the dependent libraries); should that environment change, a new performance model may need to be learned from scratch. This strong assumption limits the reusability of performance models across environments. \emph{Reusing} performance models or their byproducts across environments is demanded by many application scenarios, here we mention two common scenarios:
\begin{itemize}
\item \emph{Scenario 1: Hardware change}: The developers of a software system performed a performance benchmarking of the system in its staging environment and built a performance model. The model may not be able to provide accurate predictions for the performance of the system in the actual production environment though (\emph{e.g.}, due to the instability of measurements in its staging environment \cite{LC:TOIT,PJZ:TOIT,BV:SPEC}). 
\item \emph{Scenario 2: Workload change}: The developers of a database system built a performance model using a read-heavy workload, however, the model may not be able to provide accurate predictions once the workload changes to a write-heavy one. The reason is that if the workload changes, different functions of the software might get activated (more often) and so the non-functional behavior changes, too.
\end{itemize}


In such scenarios, not every user wants to repeat the costly process of building a new performance model to find a suitable configuration for the new environment. 
Recently, the use of transfer learning (cf. \figref{tl-general}) has been suggested to decrease the cost of learning by transferring knowledge about performance behavior across environments~\cite{JVKSK:SEAMS17,CZJ:TKDE,VPGFC:ICPE17}. Similar to humans that learn from previous experience and transfer the learning to accomplish new tasks easier, here, knowledge about performance behavior gained in one environment can be reused effectively to learn models for changed environments with a lower cost. Despite its success, it is unclear \emph{why} and \emph{when} transfer learning works for performance analysis in highly configurable systems.

To understand the \emph{why} and \emph{when}, in this paper, we conduct an exploratory empirical study, comparing performance behavior of highly configurable systems across environmental conditions (changing workload, hardware, and software versions), to explore what forms of knowledge can be commonly exploited for \textbf{performance modeling and analysis}. 
Specifically, we explore how performance measures and models across the source and target of an environmental change are related. The notion of \textbf{relatedness} across environments gives us insights to consolidate common knowledge that is \textbf{shared} implicitly between the two environments, from knowing entire \textbf{performance distributions}, to knowing about the \textbf{best or invalid configurations}, or knowing \textbf{influential configuration options}, or knowing about important \textbf{interactions}. The various forms of shared knowledge, that we discovered in this empirical study, provide opportunities to develop novel transfer learning that are not only based on correlation concept, but also more diverse forms of similarities across environments. 

More specifically, we explore several hypotheses about the notion of common knowledge across environments. Our hypotheses start with very obvious relationships (\emph{e.g.}, correlation) that can be easily exploited, but range toward more subtle relationships (\emph{e.g.}, influential options or invalid regions remain stable) that can be explored with more advanced transfer learning techniques yet to be developed.
We tested our hypotheses across \emph{36 environmental changes} in \emph{4 configurable systems} that have been selected purposefully covering different severities and varieties. For instance, we selected simple hardware changes (by changing computing capacity) as well as severe changes (by changing hardware from desktop to cloud). 

Our results indicate that some knowledge about performance behavior can be transfered even in the most severe changes we explored, and that transfer learning is actually easy for many environmental changes. 
We observed that, for small changes, we can frequently transfer performance models linearly across environments, while for severe environmental changes, we can still transfer partial knowledge, \emph{e.g.}, information about influential options or regions with invalid configurations, that can still be exploited in transfer learning, for example, to avoid certain regions when exploring a configuration space.
Overall, our results are encouraging to explore transfer learning further for building performance models, showing broad possibilities of applying transfer learning beyond the relatively small changes explored in existing work (\emph{e.g.}, small hardware changes~\cite{VPGFC:ICPE17}, low fidelity simulations~\cite{JVKSK:SEAMS17}, similar systems~\cite{CZJ:TKDE}).

Overall, our contributions are the following:
\begin{itemize}
\item We formulate a series of hypotheses to explore the presence and nature of common, transferable knowledge between a source and a target environment, ranging from easily exploitable relationships to more subtle ones.
\item We empirically investigate performance models of 4 configurable systems before and after 36 environmental changes.
We performed a thorough exploratory analysis to understand why and when transfer learning works. 
\item We discuss general implications of our results for performance modeling of configurable software systems.
\item We release the supplementary material including data of several months of performance measurements, and scripts for replication: \url{https://github.com/pooyanjamshidi/ase17}.
\end{itemize}


\section{Intuition}
\label{sec:intuition}

Understanding the performance behavior of configurable software systems can enable (i) performance debugging~\cite{SGAK:ESECFSE15,GSKA:SPPEXA}, (ii) performance tuning~\cite{H:CACM,SHM:LIO,H:AS,VPGFC:ICPE17,HPHL:ICSE15,WWHJK:GECCO15,NMSA:Arxive17,MARC:SPLC13,ORGC:SPLC14}, (iii) design-time evolution~\cite{JGAP:CSMR13,ABDD:VISSOFT13}, or (iv) runtime adaptation \cite{KK:WSR16,JVKSK:SEAMS17,HSCMAR:SIGPLAN,FHM:FSE15,EEM:TSE,EEM:FSE10}. 
A common strategy to build performance models is to use some form of sensitivity analysis~\cite{SRACCGST:Book} in which the system is executed repeatedly in different configurations and machine learning techniques are used to generalize a model that explains the influence of individual options or interactions~\cite{SGAK:ESECFSE15,VPGFC:ICPE17,GCASW:ASE13}.

In this paper, we are interested in how a performance model for a configurable system changes when we deploy the system in a different environment. To this end, we distinguish between \emph{configuration options} -- parameters that users can tweak inside the system to select functionality or make tradeoffs among performance, quality, and other attributes -- and \emph{environment changes} -- differences in how the system is deployed and used in terms of workload, hardware, and version. 
If a performance model remains relatively stable across environments (e.g., the top configurations remain the top configurations, the most influential options and interactions remain most influential), we can exploit this stability when learning performance models for new environments. Instead of building the model from scratch (as often exhaustively measuring the same configurations on a new environment), we can reuse knowledge gathered previously for other environments in a form of \emph{transfer learning}~\cite{TS:BookChapter,PY:TKDE,CZJ:TKDE}. That is, we can develop cheaper, faster and more accurate performance models that allow us to make predictions and optimizations of performance \emph{in changing environments}~\cite{JVKSK:SEAMS17}.

For example, consider an update to faster hardware. We would often expect that the system will get faster, but will do so in a nearly uniform fashion. However, we may expect that options that cause a lot of I/O operations (e.g., a backup feature) may benefit less from a faster CPU than other options; so not all environment changes will cause uniform changes. If transfer across hardware is indeed usually easy, this encourages, for example, scenarios in which we learn performance models offline on cheap hardware and transfer it to the real system with few expensive measurements for adjustment. 
The question is what kind of knowledge can be exploited across environments in practice, with simple or more advanced forms of transfer learning. Specifically, we ask whether there exists common information (i.e., \emph{transferable/reusable knowledge}, c.f., \figref{tl-general}) that applies to both source and target environments and, therefore, can be carried over across environments.



\subsection{Environmental changes}
Let us first introduce what we mean by \emph{environment}, the key concept that is used throughout this paper. 
An environmental condition for a configurable system is determined by its hardware, workload, and software version.
\emph{(i)~Hardware:} The deployment configuration in which the software system is running. 
\emph{(ii)~Workload:} The input of the system on which it operates on. 
\emph{(iii)~Version:} The state of the code base at a certain point in time. 
Of course, other environmental changes might be possible (e.g., JVM upgrade). But, we limit this study to this selection as we consider the most common changes in practice that affect performance behavior of systems.


\subsection{Preliminary concepts}

In this section, we provide definitions of concepts that we use throughout this study. The formal notations enable us to concisely convey concepts throughout the paper.

\subsubsection{Configuration and environment space}
Let $C_i$ indicate the $i$-th configuration option of a system $\mathcal{A}$, which is either enabled or disabled (the definitions easily generalize to non-boolean options with finite domains). The configuration space is a Cartesian product of all options $\mathcal{C}=Dom(C_1) \times \dots \times Dom(C_d)$, where $Dom(C_i)=\{0,1\}$ and $d$ is the number of options. A \emph{configuration} is then a member of the configuration space where all the options are either enabled or disabled. 

We describe an environmental condition $\boldsymbol{e}$ by 3 variables $\boldsymbol{e}=[h,w,v]$ drawn from a given environment space $\mathcal{E}=H \times W\times V$, where each member represents a set of possible values for the hardware $h$, workload $w$, and system version $v$.  We use notation $\boldsymbol{ec}:[h,w_1 \rightarrow w_2,v]$ as shorthand for an environment change from workload $w_1$ to workload $w_2$ where hardware and version remain stable.


\subsubsection{Performance model}
Given a software system $\mathcal{A}$ with configuration space $\mathcal{C}$ and environment space $\mathcal{E}$, a \emph{performance model} is a black-box function $f:\mathcal{C}\times \mathcal{E}\rightarrow\mathbb{R}$ that maps each configuration $\mathbf{c}\in \mathcal{C}$ of $\mathcal{A}$ in an environment $\boldsymbol{e}\in \mathcal{E}$ to the performance of the system. To construct a performance model, we run $\mathcal{A}$ in a fixed environmental condition $\boldsymbol{e}\in\mathcal{E}$ on various configurations $\mathbf{c}_i \in \mathcal{C}$, and record the resulting performance values $y_i=f(\mathbf{c}_i,\boldsymbol{e})+\epsilon_i$ where $\epsilon_i \sim \mathcal{N}(0,\sigma_i)$ is the measurement noise corresponding to a normal distribution with zero mean and variance $\sigma_i^2$. The training data for learning a performance model for system $\mathcal{A}$ in environment $\boldsymbol{e}$ is then $\mathcal{D}_{tr}=\{(\mathbf{c}_i,y_i)\}_{i=1}^n$, where $n$ is the number of measurements. 


\subsubsection{Performance distribution}
We can and will compare the performance models, but a more relax representation that allows us to assess the potentials for transfer learning is the empirical \emph{performance distribution}. The performance distribution is a stochastic process, $pd:\mathcal{E}\rightarrow\Delta(\mathbb{R})$, that defines a probability distribution over performance measures for environmental conditions of a system. 
To construct a performance distribution for a system $\mathcal{A}$ with configuration space $\mathcal{C}$, 
we fit a probability distribution to the set of performance values, $\mathcal{D}_{\boldsymbol{e}}=\{y_i\}, \boldsymbol{e}\in\mathcal{E}$, using kernel density estimation \cite{Bishop:Book} (in the same way as histograms are constructed in statistics).   
\subsubsection{Influential option} 
At the level of individual configuration options, we will be interested in exploring whether options have an influence on the performance of the system in either environment; not all options will have an impact on performance in all environments.
We introduce the notion of a \emph{influential option} to describe a configuration option that has a statistically significant influence on performance. 
\subsubsection{Options interaction} 
The performance influence of individual configuration options may not compose linearly. For example, while encryption will slow down the system due to extra computations and compression can speed up transfer over a network, combining both options may lead to surprising effects because encrypted data is less compressible.
In this work, we will look for \emph{interactions of options} as nonlinear effects where the influence of two options combined is different from the sum of their individual influences \cite{SGAK:ESECFSE15,SKKABRS:ICSE12}.

\subsubsection{Invalid configuration} We consider a configuration as invalid if it causes a failure or a timeout.

\subsection{Transferable knowledge}

As depicted in \figref{tl-general}, any sort of knowledge that can be \emph{extracted} from the source environment and can contribute to the learning of a better model (i.e., faster, cheaper, more accurate, or more reliable) in the target environment is considered as transferable knowledge (or reusable knowledge \cite{AJP:JSEP}). 
There are several pieces of knowledge we can transfer, such as (i) classification or regression models, (ii) dependency graphs that represent the dependencies among configurations, and (iii) option interactions in order to prioritize certain regions in the configuration space.
For transferring the extracted knowledge, we need a \emph{transfer function} that transforms the source model to the target model: $tf:f(\cdot,\boldsymbol{e}_s)\rightarrow f(\cdot,\boldsymbol{e}_t)$. In its simplest form, it can be a linear mapping that transforms the source model to the target: $f(\cdot,\boldsymbol{e}_t)=\alpha\times f(\cdot,\boldsymbol{e}_s)+\beta$, where $\alpha,\beta$ are learned using observations from both environments \cite{VPGFC:ICPE17}. More \emph{sophisticated transfer learning} exists that reuses source data using learners such as Gaussian Processes (GP)~\cite{JVKSK:SEAMS17}. 




\section{Research Questions and Methodology}
\label{sec:methodology}


\subsection{Research questions}
\label{sec:method}
The overall question that we explore in this paper is \emph{``why and when does transfer learning work for configurable software systems?''} Our hypothesis is that performance models in source and target environments are usually somehow ``related.'' To understand the notion of relatedness that we commonly find for environmental changes in practice, we explore several research questions (each with several hypotheses), from strong notions of relatedness (e.g., linear shift) toward weaker ones (e.g., stability of influential options): 

\emph{\textbf{RQ1}: Does the performance behavior stay consistent across environments? (\secref{rq1})}

If we can establish with RQ1 that linear changes across environments are common, this would be promising for transfer learning because even simple linear transformations can be applied. Even if not all environment changes may be amendable to this \emph{easy transfer learning}, we explore what kind of environment changes are more amendable to transfer learning than others.

\emph{\textbf{RQ2}: Is the influence of configuration options on performance consistent across environments? (\secref{rq2})}

For cases in which easy transfer learning are not possible, RQ2 concerns information that can be exploited for transfer learning at the level of individual configuration options. Specifically, we explore how commonly the influential options remain stable across environment changes.

\emph{\textbf{RQ3}: Are the interactions among configuration options preserved across environments? (\secref{rq3})} 

In addition to individual options in RQ2, RQ3 concerns interactions among options, that, as described above, can often be important for explaining the effect of performance variations across configurations. Again, we explore how commonly interactions are related across environment changes.

\emph{\textbf{RQ4}: Are the configurations that are invalid in the source environment with respect to non-functional constraints also invalid in the target environment? (\secref{rq4})}

Finally, RQ4 explores an important facet of invalid configurations: How commonly can we transfer knowledge about invalid configurations across environments? Even if we cannot transfer much structure for the performance model otherwise, transferring knowledge about configurations can guide learning in the target environment on the relevant regions.

\subsection{Methodology}
\label{sec:method}
\paragraph*{Design} 
We investigate changes of performance models across environments. Therefore, we need to establish the performance of a system and how it is affected by configuration options in multiple environments. To this end, we measure the performance of each system using standard benchmarks and repeated the measurements across a large number of configurations. We then repeat this process for several changes to the environment: using different hardware, different workloads, and different versions of the system.
Finally, we perform the analysis of relatedness by comparing the performance and how it is affected by options across environments. We perform comparison of a total of 36 environment changes.

\paragraph*{Analysis} For answering the research questions, we formulate different assumptions about the relatedness of the source and target environments as hypotheses -- from stronger to more relaxed assumptions. For each hypothesis, we define one or more metrics and analyze 36~environment changes in four subject systems described below. For each hypothesis, we discuss how commonly we identify this kind of relatedness and whether we can identify classes of changes for which this relatedness is characteristic.
If we find out that for an environmental change a stronger assumption holds, it means that a more informative knowledge is available to transfer.


\paragraph*{Severity of environment changes}\label{sec:expectedseverity}
We purposefully select environment changes for each subject system with the goal of exploring many different kinds of changes with different expected \emph{severity} of change.
With a diverse set of changes, we hope to detect patterns of environment changes that have similar characteristics with regard to relatedness of performance models.
We expect that less severe changes lead to more related performance models that are easier to exploit in transfer learning than more severe ones.
For transparency, we recorded the \emph{expected} severity of the change when selecting environments, as listed in \tabref{results}, on a scale from small change to very large change.
For example, we expect a small variation where we change the processor of the hardware to a slightly faster version, but expect a large change when we replace a local desktop computer by a virtual machine in the cloud. 
Since we are neither domain experts nor developers of our subject systems, recording the expected severity allows us to estimate how well intuitive judgments can (eventually) be made about suitability for transfer learning and it allows us to focus our discussion on surprising observations.

\subsection{Subject systems}
\label{sec:subjects}

In this study, we selected four configurable software systems from different domains, with different functionalities, and written in different programming languages (cf. \tabref{systems}).

{\sf \small SPEAR} is an industrial strength bit-vector arithmetic decision procedure and a Boolean satisfiability (SAT) solver. It is designed for proving software verification conditions and it is used for bug hunting.
We considered a configuration space with 14 options that represent heuristics for solving the problems and therefore affect the solving time. 
 We measured how long it takes to solve a SAT problem in all 16,384 configurations in multiple environments: four different SAT problems with different difficulty serve as workload, measured on three hardware system, with two versions of the solver as listed in \tabref{results}. The difficulty of the workload is characterized by the SAT problem's number of variables and clauses. 

{\sf \small x264} is a video encoder that compresses video files with a configuration space of 16 options to adjust output quality, encoder types, and encoding heuristics. Due to the size of the configuration space, we measured a subset of 4000 sampled randomly configurations. We measured the time needed to encode three different benchmark videos on two different hardware systems and for three versions as listed in \tabref{results}. Each benchmark consists of a raw video with different quality and size and we expect that options related to optimizing encoding affect the encoding time differently. We judged expected severity of environmental changes based on the difference between quality and size of benchmark videos. 

{\sf \small SQLite} is a lightweight relational database management system, embedded in several browsers and operating systems, with 14 configuration options that change indexing and features for size compression useful in embedded systems, but have performance impact. We expect that some options affect certain kinds of workload (e.g., read-heavy rather than write-heavy workloads) more than others.
We have measured 1000 randomly selected configurations on two hardware platforms for two versions of the database system; as workload, we have considered four variations of queries that focus on sequential reads, random reads, sequential write, and batch writes.

{\sf \small SaC} is a compiler for high-performance computing~\cite{SAC}. The {\sf \small SaC} compiler implements a large number of high-level and low-level optimizations to tune programs for efficient parallel executions configurable with 50 options controlling optimizations such as function inlining, constant folding, and array elimination. 
We measure the execution time of a program compiled in 71,267 randomly selected configurations to assess the performance impact of {\sf \small SaC}'s options.
As workloads, we select 10 different demo programs shipped with {\sf \small SaC}, each computationally intensive, but with different characteristics.
Workloads include Monte Carlo algorithms such as {\sf \small pfilter} with multiple optimizable loops as well as programs heavily based on matrix operations like {\sf \small srad}.

To account for measurement noise, we have measured each configuration of each system and environment 3 times and used the mean for the analyses. While many performance and quality measures can be analyzed, our primary performance metric is wall-clock execution time, which is captured differently for each systems in \tabref{systems}: execution time, encoding time, query time, and analysis time.



\begin{table}[t]
\centering
\caption{Overview of the real-world subject systems.} 
\label{tab:systems}
\resizebox{\columnwidth}{!}{
\begin{tabular}{llrrrrr}
\toprule
System            & Domain        & $d$ 	& $|\mathcal{C}|$  & $|H|$  & $|W|$  & $|V|$ \\
\midrule
{\sf SPEAR}             & SAT solver    & 14      & 16\,384	           & 3      & 4      & 2     \\
{\sf x264}              & Video encoder & 16      & 4\,000 	           & 2      & 3      & 3     \\
{\sf SQLite}            & Database      & 14      & 1\,000 	           & 2      & 14     & 2     \\
{\sf SaC} 		  	  & Compiler      & 50      & 71\,267	           & 1      & 10     & 1     \\
\bottomrule
\end{tabular}}
\begin{tablenotes}
\item {\scriptsize $d$: configuration options; $\mathcal{C}$: configurations; $H$:~hardware environments;  $W$:~analyzed workload; $V$:~analyzed versions.}
\end{tablenotes}
\end{table}

\section{Performance Behavior Consistency (RQ1)}
\label{sec:rq1}

Here, we investigate the relatedness of environments in the entire configuration space. We start by testing the strongest assumption (\emph{i.e.}, linear shift), which would enable an easy transfer learning (H1.1). We expect that the first hypothesis holds only for simple environmental changes. Therefore, we subsequently relax the hypothesis to test whether and when the performance distributions are similar (H1.2), whether the ranking of configurations (H1.3), and the top/bottom configurations (H1.4) stay consistent. \tabref{results} summaries the results.


{\noindent \textbf{H1.1}}: The relation of the source response to the target is a constant or proportional shift. 

{\noindent\textbf{Importance}}. If the target response is related to the source by a constant or proportional shift, it is trivial to understand the performance behavior for the target environment using the model that has already been learned in the source environment: We need to \emph{linearly transform} the source model to get the target model. We expect a linear shift if a central hardware device affecting the functionality of all configuration options homogeneously, changes such as the CPU, or homogeneous workload change. 
Previous studies demonstrated the existence of such cases where they trained a linear transformation to derive a target model for hardware changes~\cite{VPGFC:ICPE17}. 


{\noindent \textbf{Metric}}. We investigate whether $f(\mathbf{c},\boldsymbol{e}_t)=\alpha \times f(\mathbf{c},\boldsymbol{e}_s)+\beta, \forall\mathbf{c} \in \mathcal{C}$. We use metric \textbf{M1}: Pearson linear correlation \cite{Bishop:Book} between $f(\mathbf{c},\boldsymbol{e}_s)$ and $f(\mathbf{c},\boldsymbol{e}_t)$ to evaluate the hypothesis. If the correlation is 1, we can linearly transform performance models. Due to measurement noise, we do not expect perfect correlation, but we expect, for correlations higher than $0.9$, simple transfer learning can produce good predictions.

{\noindent\textbf{Results}}. 
The result in \tabref{results} show very high correlations for about a third of all studied environmental changes.
In particular, we observe high correlations for hardware changes and for many workload changes of low expected severity.

\textit{Hardware change}: 
Hardware changes often result in near-perfect correlations except for severe changes where we have used unstable hardware (\emph{e.g.}, Amazon cloud in $\boldsymbol{ec}_2$).
We investigated why using cloud hardware resulted in weak linear correlations. We analyzed the variance of the measurement noise and we observed that the proportion of the variance of the noise in the source to the target in $\boldsymbol{ec}_2$ is ${\bar{\sigma}^2_{\boldsymbol{ec}_2^s}}/{\bar{\sigma}^2_{\boldsymbol{ec}_2^t}}=33.39$, which is an order of magnitude larger than the corresponding one in $\boldsymbol{ec}_1$ (${\bar{\sigma}^2_{\boldsymbol{ec}_1^s}}/{\bar{\sigma}^2_{\boldsymbol{ec}_1^t}}=1.51$). This suggests that we can expect a linear transformation across environments when hardware resources execute in a stable environment. For transfer learning, this means that we could reuse measurements from cheaper or testing servers in order to predict the performance behavior~\cite{BV:SPEC}. Moreover, it also suggests that virtualization may hinder transfer learning. 

\textit{Workload change}: For {\sf \small SPEAR}, we observed very strong correlations across environments where we have considered SAT problems of different sizes and difficulties. Also, when the difference among the problem size and difficulty is closer across environments (\emph{e.g.}, $\boldsymbol{ec}_3$ vs. $\boldsymbol{ec}_4$) the correlation is slightly higher. This observation has also been confirmed for other systems. For instance, in environmental instance $\boldsymbol{ec}_3$ in {\small \sf SQLite}, where the workload change is write-heavy from sequential to batch, we have observed an almost perfect correlation, $0.96$, while in the read-heavy workload $\boldsymbol{ec}_4$ (random to sequential read) the correlation is only medium at $0.5$: First, the underlying hardware contains an SSD, which has different performance properties for reading and writing. Second, a database performs different internal functions when inserting or retrieving data. This implies that some environmental conditions may provide a better means for transfer learning. 

\textit{Version change}:
For {\sf \small SPEAR} ($\boldsymbol{ec}_{5,6,7}$) and {\sf \small x264} ($\boldsymbol{ec}_{5,6,7,8}$), the correlations are extremely weak or non existence, while for {\sf \small SQLite} ($\boldsymbol{ec}_{5}$), the correlation is almost perfect. We speculate that the optimization features that are determined by the configuration options for {\sf \small SPEAR} and {\sf \small x264} may undergo a substantial revision from version to version because algorithmic changes may significantly improve the way how the optimization features work. 
The implication for transfer learning is that code changes that substantially influence the internal logic controlled by configuration options may require a non-linear form of transformation or a complete set of new measurements in the target environment for those options only.

\begin{shaded*}
\noindent\textbf{Insight}. For non-severe hardware changes, we can linearly transfer performance models across environments. 
\end{shaded*}




{\noindent\textbf{H1.2}}: The performance distribution of the source is similar to the performance distribution of the target environment.

{\noindent\textbf{Importance}}. In the previous hypothesis, we investigated the situation whether the response functions in the source and target are linearly correlated. In this hypothesis, we consider a relaxed version of H1.1 by investigating if the performance distributions are similar. 
When the performance distributions are similar, it does not imply that there exists a linear mapping between the two responses, but, there might be a more sophisticated relationship between the two environments that can be captured by a \emph{non-linear transfer function}. 

\begin{figure}[t]
	\begin{center}
		\includegraphics[width=0.9\columnwidth]{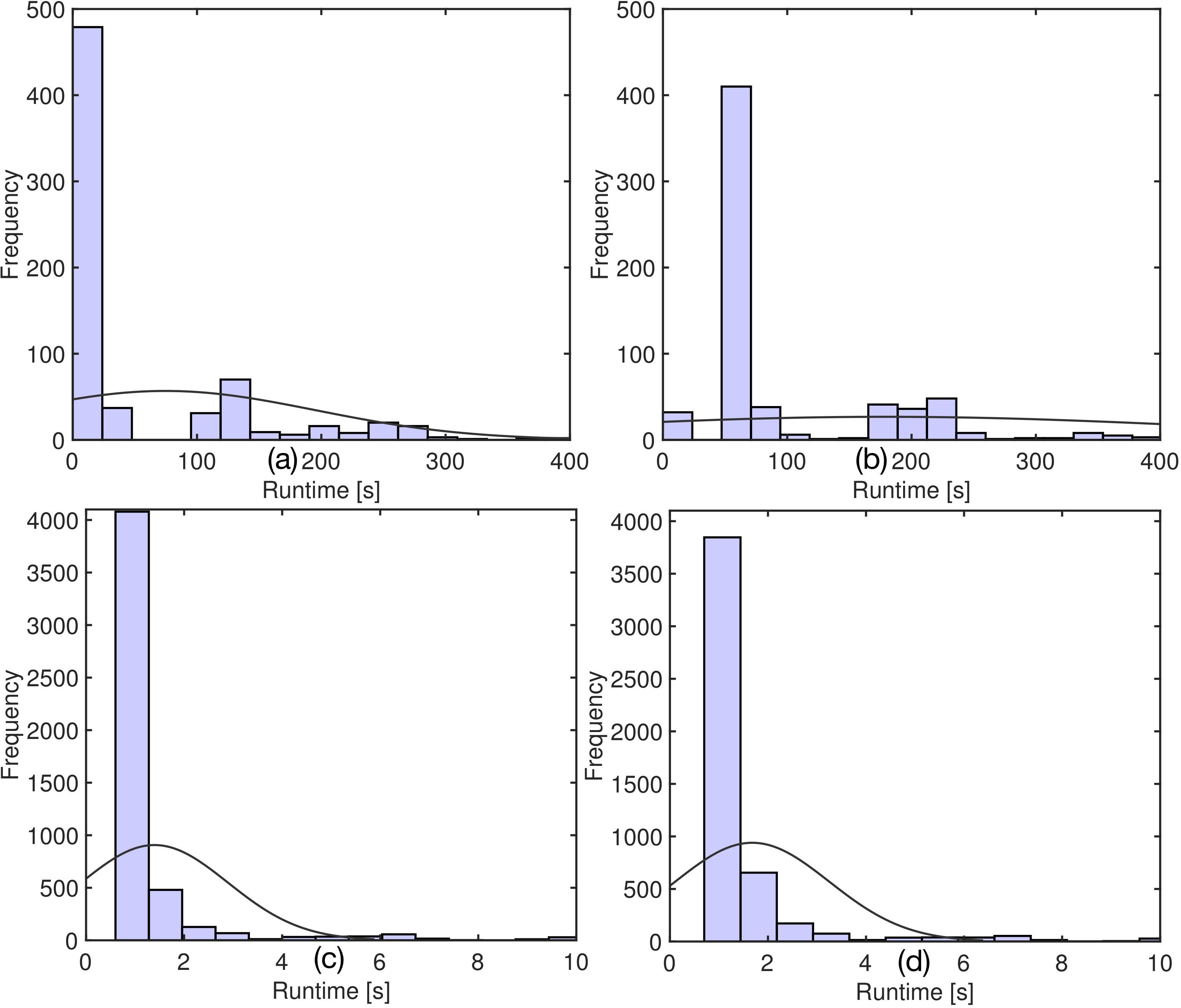}
		\caption{Performance distributions of environments can be very different, $D^{\boldsymbol{ec}_{1}}_{KL}=25.02$ (a,b), or very similar, $D^{\boldsymbol{ec}_{13}}_{KL}=0.32$ (c,d).} 
		\label{fig:performance-distributions}
	\end{center}
\end{figure}

{\noindent\textbf{Metric}}. We measure \textbf{M2}: Kullback-Leibler (KL) divergence \cite{CT:Book} to compare the similarity between the performance distributions: $D^{\boldsymbol{ec}}_{KL}(pd_s,pd_t)=\Sigma_i pd_s(\mathbf{c}_i)\log\frac{pd_s(\mathbf{c}_i)}{pd_t(\mathbf{c}_i)}$, where $pd_{s,t}(\cdot)$ are performance distributions of the source and target. As an example, we show the performance distributions of $\boldsymbol{ec}_1$ and $\boldsymbol{ec}_{13}$ and compare them using KL divergence in \figref{performance-distributions}: The lower the value of KL divergence is, the more similar are the distributions. We consider two distributions as similar if $D^{\boldsymbol{ec}}_{KL}(pd_s,pd_t)<3$~\cite{Bishop:Book} and dissimilar otherwise.

{\noindent\textbf{Results}}. 
Here, we are interested to find environmental changes for which we did not observe a strong correlation, but for which there might be similarities between the performance distributions of the environment. For $\boldsymbol{ec}_{5,6}$ in {\sf \small SPEAR}, $\boldsymbol{ec}_{3-7}$ in {\sf \small x264}, $\boldsymbol{ec}_{4,6}$ in {\sf \small SQLite}, and $\boldsymbol{ec}_{5,8}$ in {\sf \small SaC}, the performance distributions are similar across environments. This implies that there exist a possibly non-linear transfer function that we can map performance models across environments. Previous studies demonstrated the feasibility of highly non-linear kernel functions for transfer learning in configurable systems \cite{JVKSK:SEAMS17}. 



\begin{shaded*}
\noindent\textbf{Insight}. Even for some severe environmental changes with no linear correlation across performance models, the performance distributions are similar, showing the potential for learning a non-linear transfer function. 
\end{shaded*}

\noindent\textbf{H1.3}: The ranking of configurations stays stable. 

\noindent\textbf{Importance}. If the ranking of the configurations stays similar, the response function is then stable across environments. 
We can use this knowledge to prioritize certain regions in the configuration space for optimizations. 


\noindent\textbf{Metric}. Here, we use \emph{rank correlation} by measuring the \textbf{M3}:~Spearman correlation coefficient between response variables. 
Intuitively, the Spearman correlation will be high when observations have a similar rank. We consider rank correlations higher than $0.9$ as strong and suitable for transfer learning.

\noindent\textbf{Results}.
The results in \tabref{results} show that the rank correlations are high across hardware changes and small workload changes. This metric does not provide additional insights from what we have observed in H1.1. However, in one environmental change, where, due to excessive measurement noise, the linear correlation was low, $\boldsymbol{ec}_{2}$ for {\sf \small SPEAR}, the rank correlation is high. This might hint that when unstable hardware conditions exist, the overall ranking may stay stable.


\begin{shaded*}
\noindent\textbf{Insight}. The configurations retain their relative performance profile across hardware platforms. 
\end{shaded*}


\noindent\textbf{H1.4}: The top/bottom performer configurations are similar. 

\noindent\textbf{Importance}. If the top configurations are similar across environments, we can extract their characteristics and use that in the transfer learning process. For instance, we can identify the top configurations from the source and inform the optimization in the target environment~\cite{JC:MASCOTS16}. 
The bottom configurations can be used to avoid corresponding regions during sampling. Note that this is a relaxed hypothesis comparing to H1.3. 


\noindent\textbf{Metric}. We measure \textbf{M4/M5}: the percentage of (10th percentile) top/bottom configurations in the source that are also top/bottom performers in the target. 

\noindent\textbf{Results}. 
The results in \tabref{results} show that top/bottom configurations are common across hardware and small workload changes, therefore, this metric does not provide additional insights from what we have observed in H1.1.

\begin{shaded*}
\noindent\textbf{Insight}. Only hardware changes preserve top configurations across environments. 
\end{shaded*}

\section{Similarity of Influential Options (RQ2)}
\label{sec:rq2}

Here, we investigate whether the influence of individual configuration options on performance stays consistent across environments. We investigate two hypotheses about the influence strength (H2.1) and the importance of options (H2.2).


\noindent\textbf{H2.1}: The influential options on performance stay consistent. 

\noindent\textbf{Importance}. In highly dimensional spaces, not all configuration options affect the response significantly. If we observe a high percentage of common influential options across environments, we can exploit this for learning performance models by sampling across only a \emph{subset} of all configuration options, because we already know that these are the key options influencing performance. 


\noindent\textbf{Metric}. In order to investigate the option-specific effects, we use a paired t-test~\cite{Bishop:Book} to test if an option leads to any significant performance change and whether this change is similar across environments. That is, when comparing the pairs of configuration in which this option is enabled and disabled respectively, an influential option has a consistent effect to speed up or slow down the program, beyond random chance. If the test shows that an option make a difference, we then consider it as an influential option. We measure \textbf{M6/M7}: the number of influential options in source and target; We also measure \textbf{M8/M9}: the number of options that are influential in both/one environment. 



\noindent\textbf{Results}.
The results in \tabref{results} show that slightly more than half of the options, for all subject systems, are influential either in the source or target environments. From the influential options, a very high percentage are common in both. 
This can lead to a substantial reduction for performance measurements: we can fix the non-influential options and sample only along options, which we found influential from the source. 


\begin{shaded*}
\noindent\textbf{Insight}. Only a subset of options are influential which are largely preserved across all environment changes.
\end{shaded*}

\noindent\textbf{H2.2}: The importance of options stays consistent.  

\noindent\textbf{Importance}. In machine learning, each decision variable (here option) has a relative importance to predict the response and importance of the variables play a key role for in the feature selection process \cite{Bishop:Book}.
Here, we use this concept to determine the relative importance of configuration options, because in configurable systems, we face many options that if prioritized properly, it can be exploited for performance predictions \cite{JVKSK:SEAMS17}. 


\noindent\textbf{Metric}.
We use regression trees \cite{Bishop:Book} for determining the relative importance of configuration options because (i) they have been used widely for performance prediction of configurable systems \cite{GCASW:ASE13,VPGFC:ICPE17} and (ii) the tree structure can provide insights into the most essential options for prediction, because a tree splits on those options first that provide the highest information gain \cite{GCASW:ASE13}.
We derive estimates of the \emph{importance of options} for the trained trees on the source and target by examining how the prediction error will change as a result of options.
We measure \textbf{M10}:~correlation between importance of options for comparing the consistency across environments. 

\noindent\textbf{Results}.
From \tabref{results}, the correlation coefficient between the importance of options for different environmental changes is high, and the less severe a change the higher the correlation coefficients. This confirms our intuition that small changes in the environment do not affect the influence strength of an option. Some environmental changes, where the correlation were low according to M1, show a high correlation between option importance according to M10: $\boldsymbol{ec}_{6,7}$ in {\sf \small SPEAR}, $\boldsymbol{ec}_{3-7}$ in {\sf \small x264}, $\boldsymbol{ec}_{1,2,5,7-11,14}$ in {\sf \small SaC}. This observation gives further evidence that even though we did not observe a linear correlation, there might exist a non-linear relationship between performance measures. For instance, the influence of options stay the same, but interactions might change. 


\begin{shaded*}
\noindent\textbf{Insight}. The strength of the influence of configuration options is typically preserved across environments. 
\end{shaded*}

\section{Preservation of Option Interactions (RQ3)}
\label{sec:rq3}

We state two hypotheses about the preservation of option interactions (H3.1) and their importance (H3.2). 

\noindent\textbf{H3.1}: The interactions between configuration options are preserved across environments.

\noindent\textbf{Importance}. In highly dimensional configuration spaces, the possible number of interactions among options is exponential in the number of options and it is computationally infeasible to get measurements aiming at learning an exhaustive number of interactions. Prior work has shown that a very large portion of potential interactions has no influence~\cite{SKKABRS:ICSE12,KKL:Book}. 


\noindent\textbf{Metric}. One key objective here is to evaluate to what extent influential interactions will be preserved from source to target. Here, we learn step-wise linear regression models; a technique that has been used for creating performance influence model for configurable systems \cite{SGAK:ESECFSE15}. We learn all pairwise interactions, independently in the source and target environments. We then calculate the percentage of \emph{common pairwise interactions} from the model by comparing the coefficients of the pairwise interaction terms of the regression models. We concentrated on pairwise interactions, as they are the most common form of interactions~\cite{SKKABRS:ICSE12,KKL:Book}.
Similar to H2.1, we measure: \textbf{M11/M12}: The number of interactions in the source/target; 
\textbf{M13}: The number of interactions that agree on the direction of effects in the source and the target.

\noindent\textbf{Results}.
The results in \tabref{results} show three important observations: (i) only a small proportion of possible interactions have an effect on performance and so are relevant (confirming prior work); (ii) for the large environmental changes, the difference in the proportion of relevant interactions across environments is not similar, while for smaller environmental changes, the proportion is almost equal; (iii) a very large proportion of interactions is common across environments.

The mean percentage of interactions (averaged over all changes) are $25\%,28\%,10\%,6\%$ for {\sf \small SPEAR, x264, SQLite, SaC} respectively, where 100\% would mean that all pairwise combination of options have a distinct effect on performance. 
Also, the percentage of common interactions across environments is high, $96\%,81\%,85\%,72\%$ for {\sf \small SPEAR, x264, SQLite, SaC} respectively. This result points to an important transferable knowledge: interactions often stay consistent across changes. This insight can substantially reduce measurement efforts to purposefully measure specific configurations.

\begin{shaded*}
\noindent\textbf{Insight}. A low percentage of potential interactions are influential for performance model learning. 
\end{shaded*}

\noindent\textbf{H3.2}: The effects of interacting options stay similar.

\noindent\textbf{Importance}. If the effects of interacting options are similar across environments, we can prioritize regions in the configuration space based on the importance of the interactions. 


\noindent\textbf{Metric}. We measure \textbf{M14}: the correlation between the coefficients of the pairwise interaction terms in the linear model learned independently on the source and target environments using step-wise linear regression \cite{H:Stepwise}.

\noindent\textbf{Results}: 
The results in \tabref{results} reveal a very high and, in several cases, perfect correlations between interactions across environments. For several environmental changes where we previously could not find a strong evidence of transferable knowledge by previous metrics: $\boldsymbol{ec}_{8}$ in {\sf \small x264}, $\boldsymbol{ec}_{4,6,7}$ in  {\sf \small SQLite} and $\boldsymbol{ec}_{14}$ in {\sf \small SaC}, we observed very strong correlations for the interactions. The implication for transfer learning is that a linear transfer function (see H1.1) may not applicable for severe changes, while a complex transfer function may exist.



\begin{shaded*}
\noindent\textbf{Insight}. The importance of interactions is typically preserved across environments.
\end{shaded*}

\section{Invalid Configurations Similarity (RQ4)}
\label{sec:rq4}

For investigating similarity between invalid configurations across environments, we formulate two hypotheses about percentage of invalid configurations and their commonalities across environments (H4.1) and the existence of reusable knowledge that can distinguish invalid configurations (H4.2).

\noindent\textbf{H4.1}: The percentage of invalid configurations is similar across environments and this percentage is considerable.

\noindent\textbf{Importance}. If the percentage of invalid configurations is considerable in the source and target environments, this provides a motivation to carry any information about the invalid configurations across environments to avoid exploration of invalid regions and reduce measurement effort. 

\noindent\textbf{Metric}. We measure \textbf{M15/M16}: percentage of invalid configurations in the source and target, \textbf{M17}: percentage of invalid configurations, which are common between environments.

\noindent\textbf{Results}.
The results in \tabref{results} show that for {\sf \small SPEAR} and {\sf \small x264}, a considerable percentage ($\approx 50\%$) of configurations are invalid and all of them are common across environments. 
For {\sf \small SaC}, approximately {18\%} of the sampled configurations are invalid. For some workload changes the percentage of common invalid configuration is low ($\le10\%$). The reason is that some options in {\sf SaC} may have severe effects for some programs to be compiled, but have lower effects for others. 

\begin{shaded*}
\noindent\textbf{Insight}. A large percentage of configurations are typically invalid in both source and target environments. 
\end{shaded*}

\noindent\textbf{H4.2}: A classifier for distinguishing invalid from valid configurations is reusable across environments.

\noindent\textbf{Importance}. If there are common characteristics among the invalid configurations, we can learn a \emph{classifier} in the source to identify the invalid configurations and transfer the knowledge (classifier model) to the target environment to predict invalid configurations before measuring them, thus decrease cost. 


\noindent\textbf{Metric}. 
We learn a classifier using multinomial logistic regression \cite{Bishop:Book}. It is a model that is used to predict the probabilities of being invalid, given a set of configuration options. 
We measure \textbf{M18}: the correlation between the coefficients (\emph{i.e.}, the probability of the configuration being invalid) of the classification models that has been leaned independently. 

\noindent\textbf{Results}. 
The results in \tabref{results} show that for {\sf \small SPEAR} and {\sf \small x264}, the correlations between the coefficients are almost perfect. For {\sf \small SaC}, in environmental changes where the common invalid configurations are high, the correlations between coefficients are also very high. 
For two cases, $\boldsymbol{ec}_{6,7}$ in {\sf \small SPEAR}, we could not find any reusable knowledge previously with other metrics. Here, we can observe that even when the influence and interactions of all options change, the region of invalid configurations may stay the same. This means that we can avoid measurements (almost half of the space) in the target.  

\begin{shaded*}
\noindent\textbf{Insight}. Information for identifying invalid regions can be transfered, with a high confidence, across environments. 
\end{shaded*}

\section{Lessons Learned and Discussion}
\label{sec:discussion}

Based on our analyses of 36 environment changes, we can discuss lessons learned, implications and threats to validity.

\subsection{Lessons learned}


Based on the empirical results presented in this paper, we have learned that there is always some similarities that \textbf{relate} the source and target in different forms depending on the severity of the change: 
\begin{itemize}
	\item \emph{Simple changes}: We observed strong correlations between response functions (interpolating performance measures) and, therefore, there is a potential for constructing simple \textbf{linear transfer functions} across environments (\textbf{RQ1}).
	\item \emph{Large changes}: We observed very similar performance distributions (\emph{e.g.}, version changes). In these cases, we found evidence of high correlations between either options (\textbf{RQ2}) or interactions (\textbf{RQ3}) for which a non-linear transfer may be applicable. Therefore, the key elements in a performance model that has been learned on the source will not change, but the coefficients corresponding to options and their interactions might need to be relearned for the target. 
	\item \emph{Severe changes}: We have learned that a considerable part of configuration space is invalid across environmental changes that could be considered for sampling configurations in severe changes (\textbf{RQ4}).
\end{itemize}

\subsection{Implications for transfer learning research}

We provide explanations of why and when transfer learning works for performance modeling and analysis of highly configurable systems. 
While all research questions have positive answers for some environment changes and negative answers for others, as discussed above in \secref{rq1}--\secref{rq4}, the results align well with our expectations regarding the severity of change and their correspondence to the type of transferable knowledge: (i) For small environmental changes, the overall performance behavior was consistent across environments and a linear transformation of performance models provides a good approximation for the target performance behavior. (ii) For large environmental changes, we found evidence that individual influences of configuration options and interactions may stay consistent providing opportunities for a non-linear mapping between performance behavior across environments. (iii) Even for severe environmental changes, we found evidence of transferable knowledge in terms of reusability of detecting invalid from valid configurations providing opportunities for avoiding a large part of configuration space for sampling.

The fact that we could largely predict the severity of change without deep knowledge about the configuration spaces or implementations of the subject systems is encouraging in the sense that others will likely also be able to make intuitive judgments about transferability of knowledge. For example, a user of a performance analysis approach estimating low severity of an environment change can test this hypothesis quickly with a few measurements and select the right transfer learning strategy. Transfer learning approaches for easy environment changes are readily available~\cite{VPGFC:ICPE17,CZJ:TKDE,ZBN:OSR,JVKSK:SEAMS17}.

For more severe environment changes, more research is needed to exploit transferable knowledge. Our results show that that even with severe environmental change, there always is some transferable knowledge that can contribute to performance understanding of configurable systems. While some learning strategies can take existing domain knowledge into account and could benefit from knowledge about influential options and interactions~\cite{SGAK:ESECFSE15,SKKABRS:ICSE12}, it is less obvious how to effectively incorporate such knowledge into sampling strategies and how to build more effective learners based on limited transferable knowledge. While we strongly suspect that suitable transfer learning techniques can provide significant benefits even for severe environment changes, more research is needed to design and evaluate such techniques and compare to state of the art sampling and learning strategies. Specifically, we expect research opportunities regarding: 
\begin{enumerate}
\item \textbf{Sampling strategies} to exploit the relatedness of environments to select informative samples using the importance of specific regions~\cite{RP:EMNLP17} or avoiding invalid configurations. 
\item \textbf{Learning mechanisms} to exploit the relatedness across environments and learn either a linear or non-linear associations (e.g., active learning~\cite{WHS:ICML14}, domain adaptation~\cite{LCWJ:ICML15}, fine tuning a pre-trained model~\cite{GY:Arxiv}, feature transfer~\cite{YCBL:NIPS14}, or knowledge distillation~\cite{HVD:Arxiv15} in deep neural network architectures). However, efforts need to be made to make the learning less expensive. 
\item \textbf{Performance testing and debugging} of configurable systems to benefit from our findings by transferring interesting test cases covering interactions between options~\cite{TG:ECOOP16} or detecting invalid configurations~\cite{XZHZSYZP:SOSP13,XJHZLJP:OSDI16,XNLZ:CHI17}. 
\item \textbf{Performance tuning and optimization}~\cite{JC:MASCOTS16} benefit from the findings by identifying the interacting options and to perform importance sampling exploiting the importance coefficients of options and their interactions.
\item \textbf{Performance modeling}~\cite{DB:CSUR} benefit from the findings by developing techniques that exploits the shared knowledge in the modeling process, \emph{e.g.}, tuning the parameters of a queuing network model using transfer learning.   
\end{enumerate}

\subsection{Threats to validity}

\subsubsection{External validity}
We selected a diverse set of subject systems and a large number of purposefully selected environment changes, but, as usual, one has
to be careful when generalizing to other subject systems and environment changes.
We actually performed experiments with more environmental changes and with additional measurements on the same subject systems
(e.g., for {\sf \small SaC} we also measured the time it takes to compile the program not only its execution), but we excluded those
results because they were consistent with the presented data and did not provide additional insights.



\subsubsection{Internal and construct validity}
Due to the size of configuration spaces, we could only measure configurations exhaustively in one subject system
and had to rely on sampling (with substantial sampling size) for the others, which may miss effects in parts of the configuration
space that we did not sample. We did not encounter any surprisingly different observation in our exhaustively measured {\sf \small SPEAR} dataset.

We operationalized a large number of different measures through metrics. For each measure, we considered multiple alternative metrics (e.g., different ways to establish influential options) but settled usually on the simplest and most reliable metric we could identify to keep the paper accessible and within reasonable length. In addition, we only partially used statistical tests, as needed, and often compared metrics directly using more informal comparisons and some ad-hoc threshold for detecting common patterns across environments. A different operationalization may lead to different results, but since our results are consistent across a large number of measures, we do not expect any changes to the overall big picture.

For building the performance models, calculating importance of configuration options, and classifying the invalid configurations, we elected to use different machine learning models: step-wise linear regression, regression trees, and multinomial logistic regression. We chose these learner mainly because they are successful models that have been used in previous work for performance predictions of configurable systems. However, these are only few learning mechanisms out of many that may provide different accuracy and cost.

Measurement noise in benchmarks can be reduced but not avoided.
We performed benchmarks on dedicated systems and repeated each measurement 3 times.
We repeated experiments when we encountered unusually large deviations.

\section{Related Work}
\label{sec:related-work}

\subsection{Performance analysis of configurable software}

Performance modeling and analysis is a highly researched topic \cite{WFP:FOSE07}. Researches investigate what models are more suitable for predicting performance of the configurable systems, which sampling and optimization strategies can be used for tuning these models, and how to minimize the amount of measurement efforts for model training.

\emph{Sampling} strategies based on experimental design (such as Plackett-Burman) have been applied in the domain of configurable systems~\cite{SGAK:ESECFSE15,GCASW:ASE13,SGSAC:ASE15}. The aim of these sampling approaches is to ensure that we gain a high level of information from sparse sampling in high dimensional spaces.
 
\emph{Optimization} algorithms have also been applied to find optimal configurations for configurable systems:
Recursive random sampling~\cite{YK:SIGMETRICS}, hill climbing~\cite{XLRXZ:WWW04}, direct search~\cite{ZBN:OSR}, optimization via guessing~\cite{OK:SIGMETRICS07}, Bayesian optimization~\cite{JC:MASCOTS16}, and multi-objective optimization~\cite{FHM:FSE15}. The aim of optimization approaches is to find the optimal configuration in a highly dimensional space using only a limited sampling budget.
 
\emph{Machine learning} techniques, such as support-vector machines~\cite{YWLE:MASCOTS13}, decision trees~\cite{NMSA:Arxive17}, Fourier sparse functions~\cite{ZGBC:ASE15}, active learning~\cite{SGAK:ESECFSE15} and search-based optimization and evolutionary algorithms~\cite{HPHL:ICSE15,WWHJK:GECCO15} have also been used.

Our work is related to the performance analysis research mentioned above. However, we do not perform a comparison of different models, configuration optimization or sampling strategies. Instead, we concentrate on transferring performance models across hardware, workload and software version. Transfer learning, in general, is orthogonal to these approaches and can contribute to make these approaches more efficient for performance modeling and analysis. 

\subsection{Performance analysis across environmental change}


Environmental changes have been studied before. For example, in the context of MapReduce applications~\cite{YWLE:MASCOTS13}, performance-anomaly detection~\cite{SSIY:MASCOTS10}, performance prediction based on micro-benchmark measurements on different hardware~\cite{HPEGJD:PACT}, consistency-analysis of parameter dependencies~\cite{ZBN:OSR}, and performance prediction of configurable systems based on hardware variants and similarity search~\cite{TDZN:SIGMETRICS}.

Recently, transfer learning is used in systems and software engineering. For example, in the context of performance predictions in self-adaptive systems~\cite{JVKSK:SEAMS17}, configuration dependency transfer across software systems~\cite{CZJ:TKDE}, co-design exploration for embedded systems~\cite{BN:PACT16}, model transfer across hardware~\cite{VPGFC:ICPE17}, and configuration optimization~\cite{A:TR17}.
Although previous work has analyzed transfer learning in the context of select hardware changes~\cite{VPGFC:ICPE17,JVKSK:SEAMS17,CZJ:TKDE}, we more broadly empirically investigate why and when transfer learning works. That is, we provide evidence why and when other techniques are applicable for which environmental changes. 

Transfer learning has also been applied in software engineering in very different contexts, including defect predictions~\cite{KMF:ASE16,NK:FSE15,NPK:ICSE13} and effort estimation~\cite{KMM:ESE}.



\section{Conclusions}
\label{sec:conclusions}

We investigated when and why transfer learning works for performance modeling and analysis of highly configurable systems. Our results suggest that performance models are frequently related across environments regarding overall performance response, performance distributions, influential configuration options and their interactions, as well as invalid configurations.
While some environment changes allow simple linear forms of transfer learning, others have less obvious relationships but can still be exploited by transferring more nuanced
aspects of the performance model, e.g., usable for guided sampling. 
Our empirical study demonstrate the existence of diverse forms of transferable knowledge across environments that can contribute to learning faster, better, reliable, and more important, less costly performance models.

\begin{landscape}
\centering
\captionof{table}{Results indicate that there exist several forms of knowledge that can be transfered across environments and can be used in transfer learning.}
\label{tab:results}
\resizebox{0.9\columnwidth}{!}{
\begin{tabular}{llrrrrrrrrrrrrrrrrrr}  
\\
\toprule
\multicolumn{2}{c}{} & \multicolumn{5}{c}{RQ1} & \multicolumn{5}{c}{RQ2} & \multicolumn{4}{c}{RQ3} & \multicolumn{4}{c}{RQ4} \\
\cmidrule(r){3-7} \cmidrule(r){8-12}  \cmidrule(r){13-16} \cmidrule(r){17-20}
 &  & H1.1  & H1.2 & H1.3 & \multicolumn{2}{c}{H1.4}   & \multicolumn{4}{c}{H2.1} & H2.2 & \multicolumn{3}{c}{H3.1} & H3.2 & \multicolumn{3}{c}{H4.1} & H4.2   \\
\cmidrule(r){3-3} \cmidrule(r){4-4} \cmidrule(r){5-5} \cmidrule(r){6-7} \cmidrule(r){8-11} \cmidrule(r){12-12} \cmidrule(r){13-15} \cmidrule(r){16-16} \cmidrule(r){17-19} \cmidrule(r){20-20}
 Environment & ES & M1 & M2 & M3 & M4 & M5 & M6 & M7 & M8 & M9 & M10 & M11 & M12 & M13 & M14 & M15 & M16	& M17 & M18 \\
\midrule
\multicolumn{20}{l}{\cellcolor{blue!30}{\sf \small SPEAR}| {Workload} (\#variables/\#clauses): $w_1:774/5934, w_2:1008/7728, w_3:1554/11914, w_4:978/7498;$ Version: $v_1:1.2, v_2:2.7$}\\ 
\cellcolor{gray!30}$\boldsymbol{ec}_1:[h_2\rightarrow h_1,w_1,v_2]$ & S & {\bf 1.00}  & 0.22  & {\bf 0.97}  & {\bf 0.92} & {\bf 0.92} & 9  & 7 & 7 & 0 & 1 & 25 & 25 & 25 & 1.00  & 0.47 & 0.45 & 1 & 1.00 \\
\cellcolor{gray!30}$\boldsymbol{ec}_2:[h_4\rightarrow h_1,w_1,v_2]$ & L & 0.59  & 24.88 & {\bf 0.91}  & {\bf 0.76} & {\bf 0.86} & 12 & 7 & 4 & 2 & 0.51 & {\bf 41} & {\bf 27} & 21 & {\bf 0.98}  & 0.48 & 0.45 & 1 & 0.98 \\
\cellcolor{gray!30}$\boldsymbol{ec}_3:[h_1,w_1 \rightarrow w_2,v_2]$ & L & {\bf 0.96}  & 1.97  & 0.17  & 0.44 & 0.32 & 9  & 7 & 4 & 3 & 1 & 23 & 23 & 22 & 0.99  & 0.45 & 0.45 & 1 & 1.00 \\
\cellcolor{gray!30}$\boldsymbol{ec}_4:[h_1,w_1 \rightarrow w_3,v_2]$ & M & {\bf 0.90}  & 3.36  & -0.08 & 0.30 & 0.11 & 7  & 7 & 4 & 3 & 0.99 & 22 & 23 & 22 & 0.99  & 0.45 & 0.49 & 1 & 0.94\\
\cellcolor{gray!30}$\boldsymbol{ec}_5:[h_1,w_1,v_2 \rightarrow v_1]$ & S  & 0.23  & {\bf 0.30}  & 0.35  & 0.28 & 0.32 & 6  & 5 & 3 & 1 & 0.32 & {\bf 21} & {\bf 7}  & 7  & 0.33  & 0.45 & 0.50 & 1 & 0.96\\
\cellcolor{gray!30}$\boldsymbol{ec}_6:[h_1,w_1 \rightarrow w_2,v_1 \rightarrow v_2]$ & L & -0.10 & {\bf 0.72}  & -0.05 & 0.35 & 0.04 & 5  & 6 & 1 & 3 & {\bf 0.68} & {\bf 7} & {\bf 21} & 7 & 0.31  & {\bf 0.50} & {\bf 0.45} & {\bf 1} & {\bf 0.96}\\
\cellcolor{gray!30}$\boldsymbol{ec}_7:[h_1\rightarrow h_2,w_1\rightarrow w_4,v_2 \rightarrow v_1]$ & VL & -0.10 & 6.95  & 0.14  & 0.41 & 0.15 & 6  & 4 & 2 & 2 & {\bf 0.88} & {\bf 21} & {\bf 7}  & 7  & -0.44 & {\bf 0.47} & {\bf 0.50} & {\bf 1} & {\bf 0.97}\\
\addlinespace
\multicolumn{20}{l}{\cellcolor{blue!30}{\sf \small x264}| Workload (\#pictures/size): $w_1:8/2, w_2:32/11, w_3:128/44;$ Version: $v_1:r2389, v_2:r2744, v_3:r2744$} \\
\cellcolor{gray!30}$\boldsymbol{ec}_1:[h_2\rightarrow h_1,w_3,v_3]$ & SM  & {\bf 0.97} & 1.00  & {\bf 0.99} & {\bf 0.97} & {\bf 0.92} & 9  & 10 & 8  & 0 & 0.86 & 21 & 33 & 18 & 1.00 & 0.49 & 0.49 & 1 & 1\\
\cellcolor{gray!30}$\boldsymbol{ec}_2:[h_2\rightarrow h_1,w_1,v_3]$ & S  & {\bf 0.96} & 0.02  & {\bf 0.96} & {\bf 0.76} & {\bf 0.79} & 9  & 9  & 8  & 0 & 0.94 & 36 & 27 & 24 & 1.00 & 0.49 & 0.49 & 1 & 1\\
\cellcolor{gray!30}$\boldsymbol{ec}_3:[h_1,w_1 \rightarrow w_2,v_3]$ & M & 0.65 & {\bf 0.06}  & 0.63 & 0.53 & 0.58 & 9  & 11 & 8  & 1 & {\bf 0.89} & 27 & 33 & 22 & {\bf 0.96} & 0.49 & 0.49 & 1 & 1\\
\cellcolor{gray!30}$\boldsymbol{ec}_4:[h_1,w_1 \rightarrow w_3,v_3]$ & M  & 0.67 & {\bf 0.06}  & 0.64 & 0.53 & 0.56 & 9  & 10 & 7  & 1 & {\bf 0.88} & 27 & 33 & 20 & {\bf 0.96} & 0.49 & 0.49 & 1 & 1\\
\cellcolor{gray!30}$\boldsymbol{ec}_5:[h_1,w_3,v_2 \rightarrow v_3]$ & L  & 0.05 & {\bf 1.64}  & 0.44 & 0.43 & 0.42 & 12 & 10 & 10 & 0 & {\bf 0.83} & 47 & 33 & 29 & {\bf 1.00} & 0.49 & 0.49 & 1 & 1 \\
\cellcolor{gray!30}$\boldsymbol{ec}_6:[h_1,w_3,v_1 \rightarrow v_3]$ & L & 0.06 & {\bf 1.54}  & 0.43 & 0.43 & 0.37 & 11 & 10 & 9  & 0 & {\bf 0.80} & 46 & 33 & 27 & {\bf 0.99} & 0.49 & 0.49 & 1 & 1 \\
\cellcolor{gray!30}$\boldsymbol{ec}_7:[h_1,w_1 \rightarrow w_3,v_2 \rightarrow v_3]$ & L & 0.08 & {\bf 1.03}  & 0.26 & 0.25 & 0.22 & 8  & 10 & 5  & 1 & {\bf 0.78} & 33 & 33 & 20 & {\bf 0.94} & 0.49 & 0.49 & 1 & 1 \\
\cellcolor{gray!30}$\boldsymbol{ec}_8:[h_2\rightarrow h_1,w_1\rightarrow w_3,v_2 \rightarrow v_3]$ & VL & 0.09 & 14.51 & 0.26 & 0.23 & 0.25 & 8  & 9  & 5  & 2 & 0.58 & {\bf 33} & {\bf 21} & {\bf 18} & {\bf 0.94} & 0.49 & 0.49 & 1 & 1 \\
\addlinespace
\multicolumn{20}{l}{\cellcolor{blue!30}{\sf \small SQLite}| Workload: $w_1:write-seq$, $w_2:write-batch, w_3:read-rand, w_4:read-seq;$ Version: $v_1:3.7.6.3, v_2:3.19.0$}\\
\cellcolor{gray!30}$\boldsymbol{ec}_1:[h_3\rightarrow h_2,w_1,v_1]$ & S & {\bf 0.99} & 0.37 & {\bf 0.82} & 0.35 & 0.31 & 5 & 2 & 2 & 0 & 1 & 13 & 9  & 8 & 1.00 &  N/A   &   N/A   &   N/A   &  N/A \\
\cellcolor{gray!30}$\boldsymbol{ec}_2:[h_3\rightarrow h_2,w_2,v_1]$ & M & {\bf 0.97} & 1.08 & {\bf 0.88} & 0.40 & 0.49 & 5 & 5 & 4 & 0 & 1 & 10 & 11 & 9 & 1.00 &  N/A   &   N/A   &   N/A   &  N/A \\
\cellcolor{gray!30}$\boldsymbol{ec}_3:[h_2,w_1 \rightarrow w_2,v_1]$ & S  & {\bf 0.96} & 1.27 & {\bf 0.83} & 0.40 & 0.35 & 2 & 3 & 1 & 0 & 1 & 9  & 9  & 7 & 0.99 &  N/A   &   N/A   &   N/A   &  N/A \\
\cellcolor{gray!30}$\boldsymbol{ec}_4:[h_2,w_3 \rightarrow w_4,v_1]$ & M & 0.50 & {\bf 1.24} & 0.43 & 0.17 & 0.43 & 1 & 1 & 0 & 0 & 1 & 4  & 2  & 2 & {\bf 1.00} &  N/A   &   N/A   &   N/A   &  N/A\\
\cellcolor{gray!30}$\boldsymbol{ec}_5:[h_1,w_1,v_1 \rightarrow v_2]$ & M  & {\bf 0.95} & 1.00 & 0.79 & 0.24 & 0.29 & 2 & 4 & 1 & 0 & 1 & 12 & 11 & 7 & 0.99 &  N/A   &   N/A   &   N/A   &  N/A \\
\cellcolor{gray!30}$\boldsymbol{ec}_6:[h_1,w_2 \rightarrow w_1,v_1 \rightarrow v_2]$ & L  & 0.51 & {\bf 2.80} & 0.44 & 0.25 & 0.30 & 3 & 4 & 1 & 1 & 0.31 & 7  & 11 & 6 & {\bf 0.96} &  N/A   &   N/A   &   N/A   &  N/A  \\
\cellcolor{gray!30}$\boldsymbol{ec}_7:[h_2\rightarrow h_1,w_2\rightarrow w_1,v_1 \rightarrow v_2]$ & VL & 0.53 & 4.91 & 0.53 & 0.42 & 0.47 & 3 & 5 & 2 & 1 & 0.31 & 7  & 13 & 6 & {\bf 0.97} &  N/A   &   N/A   &   N/A   &  N/A \\
\addlinespace
\multicolumn{20}{l}{\cellcolor{blue!30}{\sf \small SaC}| Workload: $w_1:srad, w_2:pfilter,w_3:kmeans, w_4:hotspot, w_5:nw, w_6:nbody100, w_7:nbody150, w_8:nbody750, w_9:gc, w_{10}:cg$} \\
\cellcolor{gray!30}$\boldsymbol{ec}_1:[h_1,w_1 \rightarrow w_2,v_1]$ & L & 0.66 & 25.02 & 0.65 & 0.10 & {\bf 0.79} & 13 & 14 & 8  & 0 & {\bf 0.88} & 82  & 73  & 52  & 0.27  & 0.18 & 0.17 & 0.88 & 0.73 \\
\cellcolor{gray!30}$\boldsymbol{ec}_2:[h_1,w_1 \rightarrow w_3,v_1]$ & L & 0.44 & 15.77 & 0.42 & 0.10 & 0.65 & 13 & 10 & 8  & 0 & {\bf 0.91} & 82  & 63  & 50  & 0.56  & 0.18 & 0.12 & 0.90 & 0.84 \\
\cellcolor{gray!30}$\boldsymbol{ec}_3:[h_1,w_1 \rightarrow w_4,v_1]$ & S & {\bf 0.93} & 7.88  & {\bf 0.93} & 0.36 & {\bf 0.90} & 12 & 10 & 9  & 0 & 0.96 & 37  & 64  & 34  & 0.94  & 0.16 & 0.15 & 0.26 & 0.88 \\
\cellcolor{gray!30}$\boldsymbol{ec}_4:[h_1,w_1 \rightarrow w_5,v_1]$ & L & {\bf 0.96} & 2.82  & 0.78 & 0.06 & {\bf 0.81} & 16 & 12 & 10 & 0 & 0.94 & 34  & 58  & 25  & 0.04 & 0.15 & 0.22 & 0.19 & -0.29 \\
\cellcolor{gray!30}$\boldsymbol{ec}_5:[h_1,w_2 \rightarrow w_3,v_1]$ & M & 0.76 & {\bf 1.82}  & {\bf 0.84} & 0.67 & {\bf 0.86} & 17 & 11 & 9  & 1 & {\bf 0.95} & 79  & 61  & 47  & 0.55  & 0.27 & 0.13 & 0.83 & 0.88  \\
\cellcolor{gray!30}$\boldsymbol{ec}_6:[h_1,w_2 \rightarrow w_4,v_1]$ & S & {\bf 0.91} & 5.54  & {\bf 0.80} & 0.00 & {\bf 0.91} & 14 & 11 & 8  & 0 & 0.85 & 64  & 65  & 31  & -0.40 & 0.13 & 0.15 & 0.12 & 0.64 \\
\cellcolor{gray!30}$\boldsymbol{ec}_7:[h_1,w_2 \rightarrow w_5,v_1]$ & L & 0.68 & 25.31 & 0.57 & 0.11 & 0.71 & 14 & 14 & 8  & 0 & {\bf 0.88} & 67  & 59  & 29  & 0.05  & 0.21 & 0.22 & 0.09 & -0.13 \\
\cellcolor{gray!30}$\boldsymbol{ec}_8:[h_1,w_3 \rightarrow w_4,v_1]$ & L & 0.68 & {\bf 1.70}  & 0.56 & 0.00 & {\bf 0.91} & 14 & 13 & 9  & 1 & {\bf 0.88} & 57  & 67  & 36  & 0.34  & 0.11 & 0.14 & 0.05 & 0.67\\
\cellcolor{gray!30}$\boldsymbol{ec}_9:[h_1,w_3 \rightarrow w_5,v_1]$ & VL & 0.06 & 3.68  & 0.20 & 0.00 & 0.64 & 16 & 10 & 9  & 0 & {\bf 0.90} & 51  & 58  & 35  & -0.52 & 0.11 & 0.21 & 0.06 & -0.41\\
\cellcolor{gray!30}$\boldsymbol{ec}_{10}:[h_1,w_4 \rightarrow w_5,v_1]$ & L & 0.70 & 4.85  & 0.76 & 0.00 & {\bf 0.75} & 12 & 12 & 11 & 0 & {\bf 0.95} & 58  & 57  & 43  & 0.29  & 0.14 & 0.20 & 0.64 & -0.14\\
\cellcolor{gray!30}$\boldsymbol{ec}_{11}:[h_1,w_6 \rightarrow w_7,v_1]$ & S & 0.82 & 5.79  & 0.77 & 0.25 & {\bf 0.88} & 36 & 30 & 28 & 2 & {\bf 0.89} & 109 & 164 & 102 & {\bf 0.96}  &  N/A   &   N/A   &   N/A   &  N/A \\
\cellcolor{gray!30}$\boldsymbol{ec}_{12}:[h_1,w_6 \rightarrow w_8,v_1]$ & S & {\bf 1.00} & 0.52  & {\bf 0.92} & {\bf 0.80} & {\bf 0.97} & 38 & 30 & 22 & 6 & 0.94 & 51  & 53  & 43  & 0.99  &   N/A  &    N/A  &   N/A   &  N/A   \\
\cellcolor{gray!30}$\boldsymbol{ec}_{13}:[h_1,w_8 \rightarrow w_7,v_1]$ & S & {\bf 1.00} & 0.32  & {\bf 0.92} & 0.53 & {\bf 0.99} & 30 & 33 & 26 & 1 & 0.98 & 53  & 89  & 51  & 1.00  &   N/A  &    N/A  &  N/A    &  N/A \\
\cellcolor{gray!30}$\boldsymbol{ec}_{14}:[h_1,w_9 \rightarrow w_{10},v_1]$ & L & 0.24 & 4.85  & 0.56 & 0.44 & 0.77 & 22 & 21 & 18 & 3 & 0.69 & {\bf 237} & {\bf 226} & {\bf 94}  & {\bf 0.86}  &   N/A  &  N/A    & N/A     &   N/A  \\
\bottomrule
\end{tabular}}
\begin{tablenotes}
\footnotesize
\item ES: Expected severity of change (Sec.~\ref{sec:expectedseverity}): \textbf{S}: small change; \textbf{SM}: small medium change; \textbf{M}: medium change; \textbf{L}: large change; \textbf{VL}: very large change. 
\item {\sf \small SaC} workload descriptions: {\sf \footnotesize srad}: random matrix generator; {\sf \footnotesize pfilter}: particle filtering; {\sf \footnotesize hotspot}: heat transfer differential equations; {\sf \footnotesize k-means}: clustering; {\sf \footnotesize nw}: optimal matching; 
\item {\sf \footnotesize nbody}: simulation of dynamic systems; {\sf \footnotesize cg}: conjugate gradient; {\sf \footnotesize gc}: garbage collector. Hardware descriptions (ID: Type/CPUs/Clock (GHz)/RAM (GiB)/Disk): 
\item \textbf{h1}: NUC/4/1.30/15/SSD; \textbf{h2}: NUC/2/2.13/7/SCSI; \textbf{h3}:Station/2/2.8/3/SCSI; \textbf{h4}: Amazon/1/2.4/1/SSD; \textbf{h5}: Amazon/1/2.4/0.5/SSD; \textbf{h6}: Azure/1/2.4/3/SCSI
\item Metrics: \textbf{M1}: Pearson correlation; \textbf{M2}: Kullback-Leibler (KL) divergence; \textbf{M3}: Spearman correlation; \textbf{M4/M5}: Perc. of top/bottom conf.; \textbf{M6/M7}: Number of influential options; 
\item \textbf{M8/M9}: Number of options agree/disagree; \textbf{M10}: Correlation btw importance of options; \textbf{M11/M12}: Number of interactions; \textbf{M13}: Number of interactions agree on effects; 
\item \textbf{M14}: Correlation btw the coeffs; \textbf{M15/M16}: Perc. of invalid conf. in source/target; \textbf{M17}: Perc. of invalid conf. common btw environments; \textbf{M18}: Correlation btw coeffs
\end{tablenotes}
\end{landscape}

\section*{Acknowledgment}
This work has been supported by AFRL and DARPA (FA8750-16-2-0042). Kaestner's work is also supported by NSF awards 1318808 and 1552944 and the Science of Security Lablet (H9823014C0140). Siegmund's work is supported by the DFG under the contracts SI 2171/2 and SI 2171/3-1. We would like to thank Tim Menzies, Vivek Nair, Wei Fu, and Gabriel Ferreira for their feedback.

\bibliographystyle{abbrv}
\bibliography{bibliography.bib}

\end{document}